\pdfoutput=1

\documentclass[11pt]{article}

\usepackage[]{EMNLP2022}

\usepackage{times}
\usepackage{latexsym}
\usepackage{graphicx} 
\usepackage{float} 
\usepackage{subfigure} 
\usepackage{algorithm}
\usepackage{algorithmic}
\usepackage{amsmath}
\usepackage{booktabs}
\usepackage{multirow}
\usepackage{url}

\usepackage{amsmath}
\usepackage{accents}
\usepackage{statex}
\makeatletter
\def\wideubar{\underaccent{<!-- -->{\cc@style\underline{\mskip10mu}}}}
\def\Wideubar{\underaccent{<!-- -->{\cc@style\underline{\mskip8mu}}}}
\makeatother

\makeatletter
\def\widebar{\accentset{{\cc@style\underline{\mskip12mu}}}}
\def\Widebar{\accentset{{\cc@style\underline{\mskip14mu}}}}
\makeatother

\usepackage[T1]{fontenc}

\usepackage[utf8]{inputenc}

\usepackage{microtype}

\usepackage{inconsolata}

\title{Towards Better Document-level Relation Extraction via Iterative Inference}

\author{
    Liang Zhang\textsuperscript{1,2\footnotemark[1]}~~~
    Jinsong Su\textsuperscript{1,2\thanks{~~Equal Contribution.}}~~~
    Yidong Chen\textsuperscript{1,2\thanks{~~Corresponding Author.}}~~~
    Zhongjian Miao\textsuperscript{1,2}~~~
    Zijun Min\textsuperscript{1,2}\\ 
    \textbf{Qingguo Hu\textsuperscript{1,2}~~~
    Xiaodong Shi\textsuperscript{1,2}}\\
    \textsuperscript{1}School of Informatics, Xiamen University, China\\
    \textsuperscript{2}Key Laboratory of Digital Protection and Intelligent Processing of Intangible Cultural Heritage\\ of Fujian and Taiwan (Xiamen University), Ministry of Culture and Tourism, China\\
    \texttt{lzhang@stu.xmu.edu.cn}
    ~~~\texttt{\{jssu,ydchen\}@xmu.edu.cn}
}

\begin{document}
\maketitle
\setlength{\abovedisplayskip}{3.5pt}
\setlength{\belowdisplayskip}{3.5pt}
\begin{abstract}
Document-level relation extraction (RE) aims to extract the relations between entities from the input document that usually containing many difficultly-predicted entity pairs whose relations can only be predicted through relational inference.
Existing methods usually directly predict the relations of all entity pairs of input document in a one-pass manner, ignoring the fact that predictions of some entity pairs heavily depend on the predicted results of other pairs.
To deal with this issue, in this paper, we propose a novel document-level RE model with iterative inference.
Our model is mainly composed of two modules: 1) \textit{a base module} expected to provide preliminary relation predictions on entity pairs; 2) \textit{an inference module} introduced to refine these preliminary predictions by iteratively dealing with difficultly-predicted entity pairs depending on other pairs in an easy-to-hard manner.
Unlike previous methods which only consider feature information of entity pairs, our inference module is equipped with two \textit{Extended Cross Attention} units, allowing it to exploit both feature information and previous predictions of entity pairs during relational inference.
Furthermore, we adopt a two-stage strategy to train our model. 
At the first stage, we only train our base module.
During the second stage, we train the whole model, where contrastive learning is introduced to enhance the training of inference module.
Experimental results on three commonly-used datasets show that our model consistently outperforms other competitive baselines.
Our source code is available at \url{https://github.com/DeepLearnXMU/DocRE-II}.

\end{abstract}

\section{Introduction}
Relation extraction (RE) aims to identify the relation between two entities from raw texts. 
Due to its wide applications in many subsequent natural language processing (NLP) tasks, such as large-scale knowledge graph construction \cite{c:105} and question answering \cite{c:157}, RE has attracted increasing attention and become a fundamental NLP task.
Most of the existing works focus on sentence-level RE, where both considered entities come from a single sentence \cite{c:102,c:103}. 
However, large amounts of relations, such as relational facts from Wikipedia articles, are expressed by multiple sentences in real-world applications \cite{c:137,c:112}.
As calculated by \citet{c:104}, in the commonly-used DocRED dataset, the identifications of more than 40.7\% relational facts involve multiple sentences.
Therefore, a natural extension is document-level RE, which is required to exploit the input document to infer all relations between entities.
\setlength{\belowcaptionskip}{-10pt}
\begin{figure}[t]
\centering
\includegraphics[width=1.0 \columnwidth]{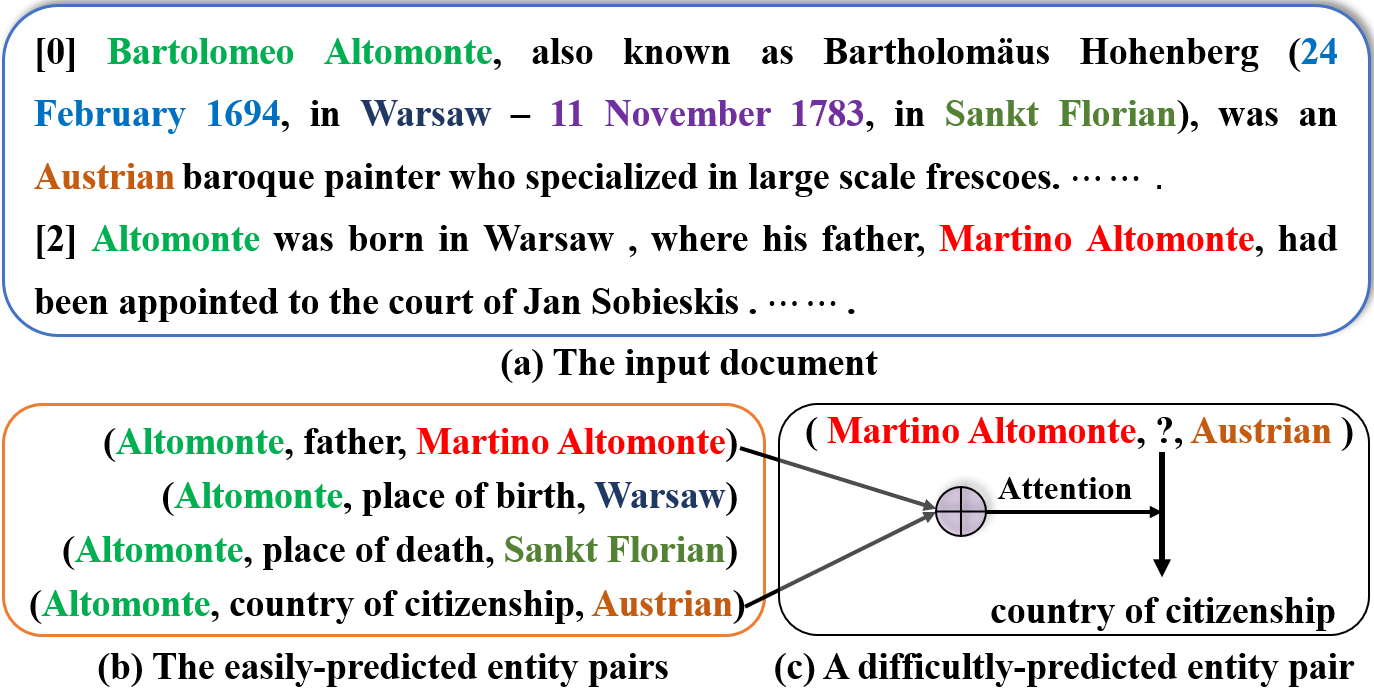} 
\caption{An example comes from the DocRED dataset. 
(a) is an input document, where different colors represent different entities. 
(b) lists some easily-predicted entity pairs whose predictions do not require reference to the predictions of other pairs.
(c) shows a difficultly-predicted entity pair, of which prediction depends on the predicted results of its overlapping pairs. 
The arrows between (b) and (c) indicate the dependencies among entity pairs, which can be exploited to benefit the relation prediction of the difficultly-predicted one.
}
\label{fig1}
\end{figure}

However, compared with sentence-level RE, document-level RE is a more challenging task.
This is because each document often contains a large number of entity pairs whose relations need to be predicted.
More importantly, the relation prediction difficulties of these pairs are usually significantly different.
The relations of some entity pairs can be directly predicted while there also exist many entity pairs whose relations can only be inferred by referring to other pairs.
As calculated by \citet{c:104}, about 61.6\% entity pairs fall into the latter category in the DocRED dataset.

Figure~\ref{fig1} shows an example from the DocRED dataset, which contains totally five entity pairs with relations.
Among these entity pairs, the relations of some pairs can be easily identified even without knowing relations of other pairs, such as (\textit{Altomonte, Martino Altomonte}) and (\textit{Altomonte, Austrian}) (See Figure~\ref{fig1}b).
By contrast, it is difficult to correctly identify the relation of (\textit{Martino Altomonte, Austrian}) because the document does not contain sufficient evidence information for it.
Furthermore, if the relations of (\textit{Altomonte, Martino Altomonte}) and (\textit{Altomonte, Austrian}) are predicted first, it will become easier to predict the relations of (\textit{Martino Altomonte, Austrian}) with the help of these previously-predicted results (See Figure~\ref{fig1}c).

To deal with the above issues, many researchers introduce graph neural networks (GNNs) \cite{c:110,c:109} to exploit the dependencies among entities or mentions for document-level RE \cite{c:107,c:105,c:106,c:108}.
However, these methods only focus on the dependencies at the entity- or mention-level while neglecting the dependencies among entity pairs, which has an important impact on the relation identification of difficultly-predicted entity pairs depending on  other pairs.
Further, \citet{c:113} model document-slevel RE as a semantic segmentation problem. 
In this way, the dependencies among entity pairs can be captured via CNN, which, however, ignores the fact that relation prediction difficulties of entity pairs are different.
Instead, \citet{c:155} treat intra- and inter-sentence entity pairs as easily-predicted and difficultly-predicted ones, and use different encoders to learn their representations, respectively.
Although this method is simple, it cannot accurately distinguish prediction difficulties of entity pairs.

In this paper, we propose a document-level RE model with iterative inference.
Overall, as shown in Figure~\ref{fig2}, our model mainly consists of two modules: 
1) \textit{an base module} used to firstly predict the relations of entity pairs roughly; 
2) \textit{a inference module} exploiting the prediction results of the previous iteration to refine the predictions in an iterative manner.
Particularly, we equip the inference module with an attention mechanism, which enables the module to accurately exploit the dependencies among overlapping entity pairs.
By doing so, we expect that based on the prediction results of the previous iteration, inference module can deal with difficultly-predicted entity pairs depending on other pairs in an easy-to-hard manner.

Furthermore, we adopt a two-stage strategy to train our model. 
First, we only train our base module.
Second, we train the entire model, where the base module is set with a relatively small learning rate to keep its parameters and performance stable.
Particularly, at the second stage, we introduce contrastive learning to encourage inference module to exploit the previously-predicted results better.

To investigate the effectiveness of our model, we conduct several groups of experiments on commonly-used datasets.
Experimental results and in-depth analyses strongly demonstrate the superiority of our model. 


\section{Our Proposed Model}
\begin{figure}[t]
\centering
\includegraphics[width=0.93 \columnwidth]{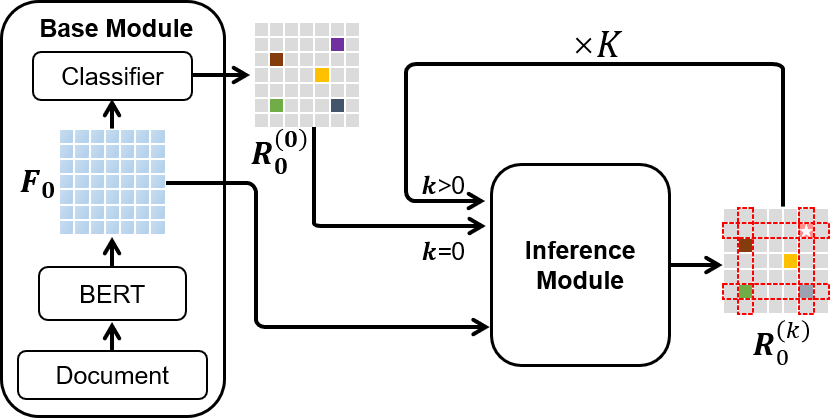}
\caption{
The overall architecture of our model.
We first use base module to make preliminary predictions.
Then, inference module gradually revises the preliminary predictions through $K$-iteration inference.
$F_0$ and $R^{(*)}_0$ refer to the feature matrix and relation matrix of entity pairs, respectively.
Note that the initial input of inference module is ($F_0{,}R^{(0)}_0$) ($k$=0), while the input of inference module becomes ($F_0{,}R^{(k)}_0$) in the subsequent iterations ($k$>0).
}
\label{fig2}
\end{figure}

In this section, we elaborate on our model. 
As shown in Figure~\ref{fig2}, our model is composed of two modules: base module (Section \ref{sec2.1}) and inference module (Section \ref{sec2.2}).
Finally, we give a detailed description of the model training (Section \ref{sec2.3}).

\subsection{Base Module}
\label{sec2.1}
We chose a competitive baseline model, ATLOP \cite{c:112}, to construct our base module.
It is used to make preliminary predictions on relations of entity pairs providing basic information for inference module.
Here, we give a brief introduction to ATLOP.
Please refer to \cite{c:112} for more details.

Let $D\!=\![w_1,w_2,...,w_L]$ denotes the input document that contains a set of entities $\{e_i\}^{N}_{i=1}$.
Note that an entity $e_i$ may appear multiple times in the document by mentions $\{m^i_j\}^{N_{e_i}}_{j=1}$.
To obtain better entity representations, we first insert a special symbol “$*$” at the start and end of the mention to mark its span.
Then, we feed the document into a pre-trained language model, obtaining its contextual embeddings: $H{=}[h_1,h_2,...,h_L]$.
Particularly, we take the context embedding of each mention's start symbol “$*$” as its feature vector.
Finally, via \textit{logsumexp} pooling \cite{c:120}, we aggregate all mention-level context embeddings of entity $e_i$ to obtain its final representation vector, \textit{i.e.}, 
$h(e_i)=\log\sum_{j=1}^{N_{e_i}} \rm{exp} \it(h(m_j^i))$.

Unlike ATLOP directly using a bilinear function to predict the relations of entity pairs, we firstly compute a feature vector for each entity pair, and then stack a single linear layer to predict their relations.
Specifically, for an entity pair ($e_s,e_o$), we calculate a local context vector $c_{s,o}$ according to the attention matrix $A$ of the last BERT layer:
\begin{align}
    c_{s,o}&= H^\mathsf{T} \frac{A_s \circ A_o}{\mathbf{1}^\mathsf{T}(A_s \circ A_o)}, 
\end{align}
where $A_s$ and $A_o$ denote the attention weights of entity $e_s$ and $e_o$ to all tokens in document, respectively, and $\circ$ refers to element-wise multiplication.
By doing so, $c_{s,o}$ can effectively capture the local contextual information related to both entities, which can be further used to enhance the representation of entity pair $(e_s, e_o)$.
Afterwards, we compute the initial feature vector $F_{s,o}$ of ($e_s{,}e_o$) as
\begin{align}
    F_{s,o}=\rm{FNN}\it \big(\big[&\tanh(W_s[h({e_s}),c_{s,o}]), \\ &\tanh(W_o[h({e_o}),c_{s,o}])\big) , 
\end{align}
where $\rm{FNN(\cdot)}$ refers to a feed-forward neural network, $W_o$ and $W_s$ are learnable weight matrices.
Finally, we obtain the probability distribution $p_{s,o}$ of relations assigned to ($e_s,e_o$) through a simple linear layer:
\begin{align}
    p_{s,o}=\sigma(W_r F_{s,o}+b_r), \label{F:4}
\end{align}
where $W_r$ and $b_r$ are model parameters.

\subsection{Inference Module}
\label{sec2.2}
\begin{figure}[t]
\setlength{\abovecaptionskip}{0.1cm}
\centering
\includegraphics[width=0.95 \columnwidth]{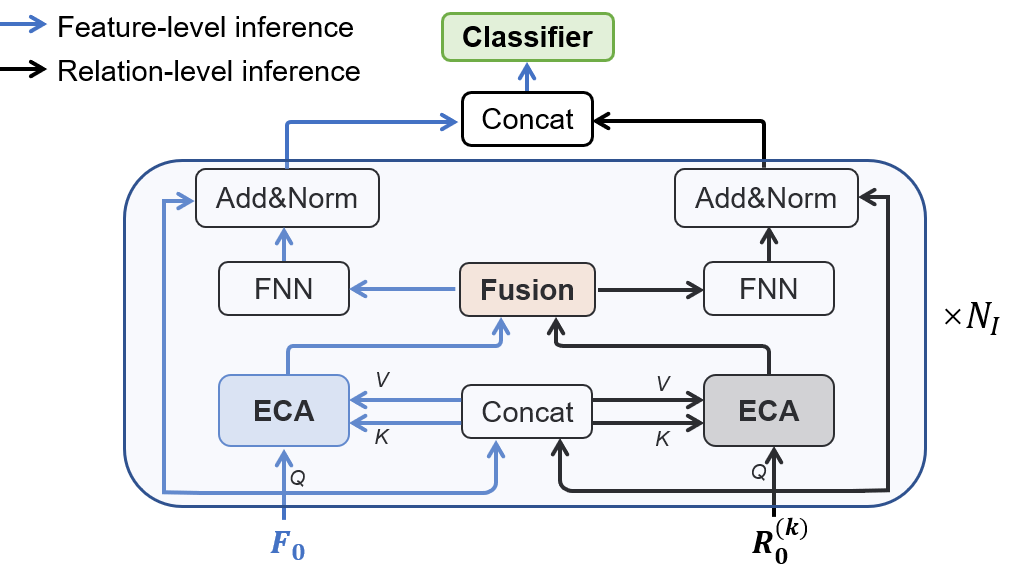} 
\caption{
Illustration of Inference Module.
It contains $N_I$ inference layers and a classifier.
With two ECA units, inference layers perform feature- and relation-level inference on the feature and relation matrices.
}
\label{fig3}
\end{figure}

As mentioned previously, we introduce inference module to iteratively refine the predictions of base module, until the maximal iteration number $K$ is reached.
As illustrated in Figure~\ref{fig3}, this module consists of $N_I$ inference layers and a classifier. 
Inference layers are used to perform inference, where the dependencies among overlapping entity pairs are leveraged to learn better entity pair representations.
Then, with the output of the top inference layer, the classifier produces better predictions.

To facilitate the computation of inference module, we combine the feature vectors of all entity pairs into a \textit{feature matrix} $F_0{=}[F_{s,o}]_{N \times N}$, where each row $F_{0[s{,}*]}$ corresponds to a subject entity $e_s$ and each column $F_{0[*{,}o]}$ corresponds to an object entity $e_o$.
Similarly, we employ the embedding operation $\rm{emb}\it(\cdot)$ to construct a \textit{relation matrix} $R^{(0)}_0{=}[\rm{emb}\it({\operatorname{arg\,max} }\,(p_{s,o}))]_{N \times N}$ from the prediction results of base module, where $p_{s,o}$ is calculated in Equation (\ref{F:4}), the subscript $*$ and superscript $\rm(*)$ of $R^{(0)}_0$ denote the inference layer index and iterative index, respectively.

\textbf{Inference Layer}
At the $k$-th iteration, with $F_0$ and $R^{(k)}_0$ as inputs, inference layers are committed to learning more expressive feature matrix $F_{N_I}$ and relation matrix $R^{(k)}_{N_I}$.
Back to Figure~\ref{fig3}, each inference layer contains three core components: 1) two \textit{Extended Cross Attention} (ECA) units performing feature- and relation-level inference, respectively (See the blue and black lines in Figure~\ref{fig3}), and 2) a \textit{Fusion sub-layer} combining the outputs of ECA units. 
Since inference layers perform the feature- and relation-level inference in the same way, we take the feature-level inference as an example to illustrate its details.

ECA is a variant of the conventional multi-head self-attention.
The basic intuition behind ECA is that for each entity pair ($e_s,e_o$), its overlapping entity pairs can provide important information for inferring its relations.
To model this intuition, we extend the multi-head self-attention to ECA that only focuses on overlapping entity pairs.
Note that these overlapping entity pairs are only located in the $s$-th row, $s$-th column, $o$-th row and $o$-th column of the feature matrix.
To effectively exploit their information, we equip ECA with four attention heads to capture their effects, respectively.
For example, at the $l$-th layer, we calculate the first attention head of ECA as follows:
\begin{align}
    \rm{head}\it_{1}{=}\rm{Attention}\it(&F_{l[s{,}o]} ,M_{l[s{,}*]} ,M_{l[s{,}*]}) , \\
    \rm{Attention}(Q,K,V) &= \rm{softmax}(\frac{QK^\mathsf{T}}{\sqrt{d_k}})V,
\end{align}
where $M_l{=}[F_l,R^{(k)}_l]$ is the concatenated matrix of $F_l$ and $R^{(k)}_l$, $M_{l[s{,}*]}$ denotes the $s$-th row of $M_l$, $d_k$ is the dimension of key ($K$) and query ($Q$).
Then, we merge all attention heads to obtain the output of ECA, which is a temporary feature matrix $\widetilde{F}_l$:
\begin{equation}
	\begin{split}
    	\widetilde{F}_l = \rm{Concat}(\rm{head}_{1},...,\rm{head}_{4})W_O,\\
	\end{split}
\end{equation}
where $W_O$ is a model parameter matrix.
Likewise, we use the other ECA to obtain a temporary relation matrix $\widetilde{R}^{(k)}_l$.
Subsequently, fusion sub-layer combines $\widetilde{F}_l$ and $\widetilde{R}^{(k)}_l$ through a gating mechanism:
\begin{align}
    s_l &= \rm{Sigmoid}\it([\widetilde{F}_l,\widetilde{R}^{(k)}_l]W_g+b_g),\\
	\widetilde{M}_l &= s_l\circ\widetilde{F}_l+(1-s_l)\circ\widetilde{R}^{(k)}_l,
\end{align}
where $W_g$ and $b_g$ are trainable parameters.

Finally, we obtain the output of the $l$-th inference layer as follows:
\begin{align}
    F_{l+1} &{=} \rm{LayerNorm}\it(F_l+\rm{FNN}\it(\widetilde{M}_l)),\\
    R^{(k)}_{l+1} &{=} \rm{LayerNorm}\it(R^{(k)}_l+\rm{FNN}\it(\widetilde{M}_l)),
\end{align}
where $\rm{LayerNorm}\it(\cdot)$ is layer normalization \cite{c:162}. 

After repeating this inference process $N_I$ times, we get the updated feature matrix $F_{N_I}$ and relation matrix $R^{(k)}_{N_I}$.

\textbf{Classifier}
On the basis of $R^{(k)}_{N_I}$ and $F_{N_I}$, we stack a single-layer classifier to provide more refined predictions on entity pairs:
\begin{align}
    P^{(k+1)}&=\sigma(W_c \big[F_{N_I},R^{(k)}_{N_I}\big]+b_c). \label{F:12}
\end{align}
Finally, we obtain the relation matrix of the ($k$+$1$)-th iteration: $R^{(k+1)}_0=\rm{emb}\it \big(\mathop{\arg\max}(P^{(k \rm {+}1)})\big)$.

We perform the above-mentioned inference for $K$ times, obtaining the final predictions $P^{(K)}$.

\subsection{Model Training}
\label{sec2.3}
We adopt a two-stage strategy to train our model. 
At the first stage, we use an \textit{adaptive threshold loss} $\mathcal{L_R}$ to only train base module.
At the second stage, we train the entire model, where we set a relatively small learning rate for base module to maintain its parameters and performance stable.
Particularly, in addition to $\mathcal{L_R}$, we introduce a \textit{contrastive loss} $\mathcal{L_C}$ to enhance the training of inference module.
To improve the training efficiency at this stage, following \citet{c:156}, we train inference module to directly correct the predictions of base module in a one-pass manner, as opposed to the multiple iterations used during testing.
Next, we describe our losses in detail.

\begin{figure}[t]
\centering
\includegraphics[width=1.0 \columnwidth]{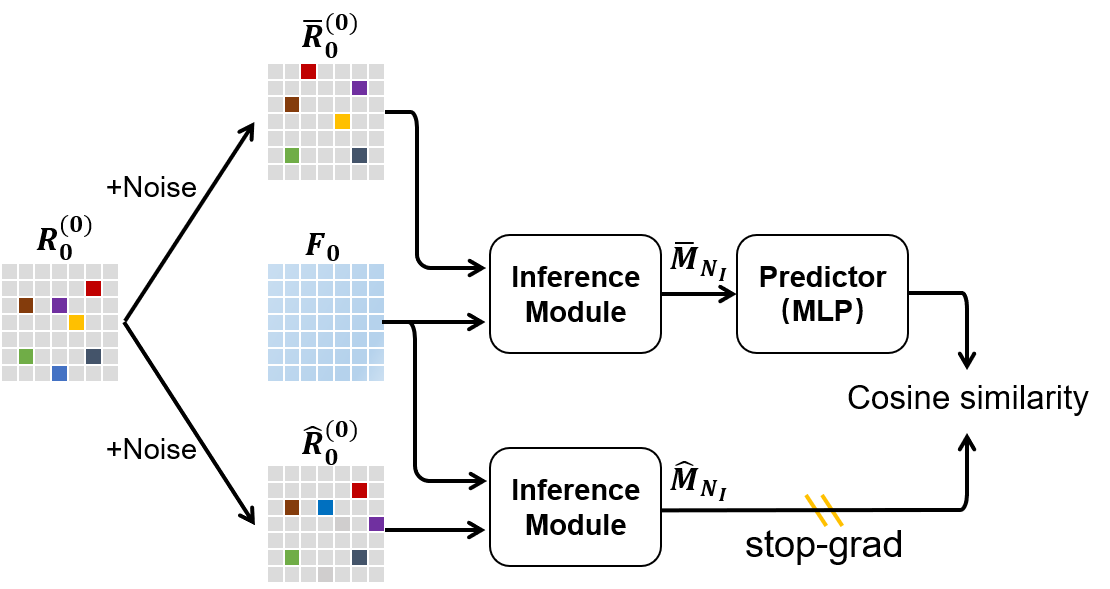} 
\caption{
Illustration of our contrastive learning.
The stop-gradient operation \textit{stop-grad} can effectively prevent the representation space of the model from collapsing.
Note the feature matrix and relation matrix fed into the two inference modules sharing parameters are in reverse order, which forces the inference module to treat both types of matrices equally.
}
\label{fig4}
\end{figure}

\textbf{Adaptive Threshold Loss $\mathcal{L_R}$}
This loss is proposed by \citet{c:112}, aiming to alleviate the imbalanced relation distribution problem in document-level RE.
In this work, we introduce this loss into our model training, which is defined as
\begin{align}
    \mathcal{L_R}{=}{-}&\sum_{r{\in}\mathcal{P}_T} \log \Big(\frac{\exp(\rm{logit}_{\it{r}})}{\sum_{r'{\in}\{\mathcal{P}_T,\rm{TH}\}} \exp(\rm{logit}_{\it{r'}})}\Big)  \nonumber \\
    {-}&\log \Big(\frac{\exp(\rm{logit}_{TH})}{\sum_{r'{\in}\{\mathcal{N}_T,\rm{TH}\}}\exp(\rm{logit}_{\it{r'}})}\Big),
\end{align}
where $\rm TH$ is a threshold relation used to distinguish between positive and negative relations.
This loss will push the logits of all positive relations $\mathcal{P}_T$ to be higher than that of TH, and pull the logits of all negative relations $\mathcal{N}_T$ to be lower than that of TH.


\textbf{Contrastive Loss $\mathcal{L_C}$}
To prevent inference module from just simply replicating the predictions of base module, we inject noises into the relation matrix $R^{(0)}_0$ by randomly substituting $r$ percent of predicted relations with incorrect ones.
However, such a noises injection renders the relation matrix $R^{(0)}_0$ less stable than the feature matrix $F_0$.
As a result, inference module prefers predicting relations using only $F_0$.
To address this defect, we introduce a variant method of contrastive learning, SimSiam \cite{c:165}, to enhance the training of our inference module.
Unlike the conventional contrastive learning, such as \cite{c:167,c:168,c:170}, it only focuses on pulling together the representations of examples in \textit{positive pairs}.
In this way, it can avoid the drawback of pushing \textit{negative pairs} far apart in conventional contrastive learning, which may limit the potential of inference module to capture the dependencies among entity pairs.


Concretely, as shown in Figure~\ref{fig4}, we generate two different relation matrices, $\overline{R}^{(0)}_0$ and $\widehat{R}^{(0)}_0$, by adding various noises to relation matrix $R^{(0)}_0$.
Then, we obtain a positive pair consisting of two examples $(F_0,\overline{R}^{(0)}_0)$ and $(\widehat{R}^{(0)}_0,F_0)$, which are fed into inference module to produce the corresponding outputs $\overline{M}_{N_I}$ and $\widehat{M}_{N_I}$, respectively.
Please note that the input order of the feature and relation matrices is inverted in these two examples.
Finally, we define the contrastive loss as follows:
\begin{equation}
	\begin{split}
       \mathcal{L_C} {=} 2 {-} \Big( \rm{Cosine}\big(\rm{MLP}(\it \widehat{M}_{N_I}),\rm{SG}(\it \overline{M}_{N_I} \big)\\
       +\rm{Cosine}\big(\rm{MLP}(\it \overline{M}_{N_I}), \rm{SG} (\it\widehat{M}_{N_I}) \big)\Big),\\
	\end{split}
\end{equation}
where $\overline{M}_{N_I}{=}[F_{N_I},\overline{R}^{(0)}_{N_I}]$ and $\widehat{M}_{N_I}{=}[F_{N_I},\widehat{R}^{(0)}_{N_I}]$ are the outputs of the top inference layer, $\rm{Cosine}(\cdot)$ indicates a cosine similarity function, $\rm{SG}(\cdot)$ refers to a stop-gradient operation that prevents the model training from collapsing \cite{c:165}, and $\rm{MLP}(\cdot)$ stands for a multi-layer perceptron function, our predictor, which helps the model learn better representations \cite{c:165,c:167}.
In this way, inference module will be encouraged to fully utilize the relation matrix.


\section{Experiments}
\subsection{Datasets}
\textbf{DocRED} \cite{c:104} 
This dataset is a large-scale crowdsourced dataset for document-level RE, which is constructed from Wikipedia and Wikidata. 
It contains 3,053 documents for training, 1,000 for development, and 1,000 for testing.
Each document contains 26 entities on average. 
In total, this dataset involves 97 target relations.

\textbf{CDR} \cite{c:122} 
It is a biomedical dataset that is constructed from the PubMed abstracts.
CDR has only one target relation: \textit{Chemical-Induced-Disease} between chemical and disease entities.
It includes about 1,500 human-annotated documents, which are equally split into training, development and test sets.

\textbf{GDA} \cite{c:123} 
It is also a biomedical dataset, which is constructed from MEDLINE abstracts via distant supervision.
Following \citet{c:107}, we split its training set into two parts: training set (23,353 documents) and development set (5,839 documents), and directly evaluate the model on its test set (1,000 documents).


\begin{table*}[th]
\centering
\small
{
\begin{tabular}{lcccccc}
\toprule
Model            & \multicolumn{4}{c}{Dev}                  & \multicolumn{2}{c}{Test} \\
                        \cmidrule(r){2-5}  \cmidrule(r){6-7}
                     & Ign$F_1$         & $F_1$     & Intra-$F1$    & Inter-$F1$   & Ign$F_1$     & $F_1$  \\ [2pt]\toprule
GEDA-BERT \cite{c:130}       & 54.52    & 56.16     & $-$             & $-$       & 53.71      & 55.74       \\
LSR-BERT \cite{c:106}        & 52.43    & 59.00        &65.26          &52.05    & 56.97      & 59.05       \\
GLRE-BERT \cite{c:108}       & $-$        & $-$         & $-$             & $-$       & 55.40       & 57.40        \\
GAIN-BERT \cite{c:105}       & 59.14    & 61.22     &67.10          &53.90     & 59.00         & 61.24       \\ 
HeterGSAN-BERT \cite{c:134}  & 58.13    & 60.18     & $-$             & $-$        & 57.12      & 59.45       \\ 
SSAN-BERT \cite{c:134}       & 56.68    & 58.95     &$-$          &$-$     & 56.06         & 58.41       \\ \midrule
BERT \cite{c:135}            & $-$        & 54.16     &61.61          &47.15      & $-$          & 53.20        \\
BERT-Two-Step \cite{c:111}   & $-$        & 54.42     &61.80          &47.28     & $-$          & 53.92       \\
HIN-BERT \cite{c:111}         & 54.29       & 56.31     & $-$             & $-$         & 53.70       & 55.60        \\
CorefBERT \cite{c:146}        & 55.32       & 57.51     & $-$             & $-$         & 54.54      & 56.96       \\
ATLOP-BERT \cite{c:112}       & 59.22       & 61.09     & $-$             & $-$         & 59.31      & 61.30       \\ \midrule
DocuNet-BERT \cite{c:113}     & 59.86       & 61.83     & $-$             & $-$    & 59.93      & 61.86  \\
SIRE-BERT \cite{c:155}        & 59.82     & 61.60     & 68.07         & 54.01     & 60.18     & 62.05 \\  
KD-BERT \cite{c:159}        & 60.08     & 62.03     & $-$         & $-$     & 60.04     & 62.08 \\ \midrule
Ours-BERT         & \textbf{60.75}$\pm$0.12 & \textbf{62.74}$\pm$0.15 & \textbf{69.14}$\pm$0.10 & \textbf{55.54}$\pm$0.19 & \textbf{60.68} & \textbf{62.65} \\ \bottomrule
\end{tabular}
}
\caption{\label{tab1}
The model performance on the development and test sets of DocRED. 
We run experiments 5 times with different random seeds and report the mean and standard deviation on the development set.
We save the best checkpoint on the development set and then report the official test scores on the CodaLab scoreboard.
The results of RoBERTa-large-based model are reported in Appendix~\ref{appendix-c}.
}
\end{table*}

\subsection{Settings}
We develop the proposed model based on PyTorch. We used BERT-base \cite{c:125}  or RoBERTa-large \cite{c:126} as the encoder on DocRED and SciBERT \cite{c:127} on CDR and GDA.
Inspired by \citet{c:156}, we sample the noise rate $r$ of the relation matrix from a uniform distribution $U(0, 0.4)$ during training.
We apply AdamW \cite{c:128} to optimize our model, with a linear warmup \cite{c:129} during the first 6\% steps followed by a linear decay to 0. 
All hyper-parameters are tuned on the development set, and some of them are listed in Appendix~\ref{appendix-b}.

\subsection{Baselines}
We compare our model with the following two types of baselines:

\textbf{Graph-based Models}
These models first construct a document graph from the input document and then perform inference on the graph through GNNs.
We include EoG \cite{c:107}, DHG \cite{c:139}, GEDA \cite{c:130}, LSR \cite{c:106}, GLRE \cite{c:108}, GAIN \cite{c:105}, HeterGSAN \cite{c:134}, and SSAN \cite{c:134} for comparison.

\textbf{Transformer-based Models}
These models directly use the pre-trained language models for document-level RE without graph structures. We compare BERT-base \cite{c:135}, BERT-TS \cite{c:135}, HIN-BERT \cite{c:111}, Coref-BERT \cite{c:146}, and ATLOP-BERT \cite{c:112} with our model.
    
Besides, we consider some recently-proposed methods that exploit the dependencies among entity pairs to improve the performance of document-level RE, including DocuNet \cite{c:113}, SIRE \cite{c:155}, and KD \cite{c:159}.

\begin{table}[t]
\centering
\small 
{
\begin{tabular}{lcccccc}
\toprule
$K$/$N_I$                           & 1         & 2             & 3         & 4       \\ [2pt]\toprule
1    & 61.53     & 61.85         & 61.80     & 61.66    \\
2    & 61.96     & 62.29         & 62.07     & 62.01     \\ 
3    & 62.41     & 62.74         & 62.66     & 62.47     \\ 
4    & 62.25     & 62.65         & 62.53     & 62.33      \\ \bottomrule
\end{tabular}
}
\caption{\label{tab8}
The performance ($F_1$ points) of our model with different values of $K$ and $N_I$ on the development set of DocRED.
}
\end{table}

\subsection{Effect of Iteration Number $K$ and Layer Number $N_I$}
The iteration number $K$ of inference module and the layer number $N_I$ of inference layers are two important hyper-parameters of our model, which directly affect the performance of inference module. 
Thus, we conduct an experiment with different values of $K$ and $N_I$ on the development set of DocRED.
From Figure~\ref{fig5}, we observe that our model achieves the best performance when $K$ and $N_I$ are set to 3 and 2, respectively.
Hence, we use $K$=3 and $N_I$=2 in all subsequent experiments.

\subsection{Main Results}
\textbf{Results on DocRED}
Following \citet{c:105}, we use $F_1$ and Ign$F_1$ as the evaluation metrics. 
Ign$F_1$ denotes the $F1$ points excluding the relational facts that are shared by the training and developmen/test sets. 
As shown in Table~\ref{tab1}, our model consistently outperforms all baselines.
Besides, we draw the following interesting conclusions:

First, our model performs better than ATLOP-BERT, our base model, by \textbf{1.35} $F_1$ and \textbf{1.37} Ign$F_1$ points on the test set, which demonstrates the effectiveness of our inference module.

Second, our model also obtains improvements of \textbf{1.41} $F_1$ and \textbf{1.86} Ign$F_1$ points on the test set, compared with the graph-based SOTA model, GAIN-BERT, which exploits the entity- and mention-level dependencies for document-level RE.
These results demonstrate that the dependencies among entity pairs are more important for document-level RE than ones among entities or mentions.

Third, our model surpasses SIRE-BERT and KD-BERT, both of which use an attention mechanism to capture the dependencies among entity pairs.
This verifies that our model can more effectively capture the dependencies among entity pairs.

Finally, we follow \citet{c:105,c:155} to report Intra-$F_1$ and Inter-$F_1$ points in Table~\ref{tab1}.
Please note that these two metrics only consider intra- and inter-sentence relations, respectively. 
Compared with Intra-$F_1$, Inter-$F_1$ can better reflect the inference ability of the model.
In terms of Inter-$F_1$, our model surpasses SIRE-BERT by \textbf{1.53} points.


\textbf{Results on the Biomedical Datasets}
We also conduct experiments on the biomedical datasets, of which results are shown in Table~\ref{tab2}.
Our model still consistently outperforms all previous baselines.
On CDR and GDA, our model obtains $F_1$ points of \textbf{73.2} and \textbf{85.9}, with absolute improvements of \textbf{3.8} and \textbf{2.0} over our base model (ATLOP-SciBERT), respectively.
Thus, we confirm that our model is also genralized to biomedicine.

\begin{table}[]
\centering
\small
{
\begin{tabular}{lcc} 
\toprule
Model               & CDR          & GDA         \\ [2pt] \toprule
BRAN \citep{c:137}                & 62.1         & $-$           \\
EoG \citep{c:107}               & 63.6         & 81.5        \\
LSR \citep{c:106}                & 64.8         & 82.2        \\
DHG \citep{c:139}                & 65.9         & 83.1        \\
SciBERT \citep{c:127}        & 65.1         & 82.5        \\
ATLOP-SciBERT \cite{c:112}   & 69.4         & 83.9        \\  
SIRE-BioBERT \cite{c:155}   & 70.8         & 84.7        \\ [2pt] \toprule
Ours-SciBERT         & \textbf{73.2} & \textbf{85.9}    \\
\bottomrule
\end{tabular}
}
\caption{\label{tab2}
The $F_1$ points on the CDR and GDA test sets.
}
\end{table}

\subsection{Ablation Study}
To investigate the effectiveness of different components on our model, we further compare our model with the following variants in Table~\ref{tab3}. 

(1) \textit{w/o fusion sub-layer}.
In this variant, we remove the fusion sub-layer from inference module, which leads to a drop of \textbf{0.37} $F_1$ points.
It suggests that combining the feature- and relation-level inference information of entity pairs is indeed useful for improving the performance of the model.

(2) \textit{w/o ECA}.
In this variant, we replace each ECA unit with a standard multi-head self-attention, where all other entity pairs can be considered.
This change causes a significant performance decline.
The underlying reason is that focusing on all other entity pairs introduces many noises to our model.

\begin{figure}[t]
\centering
\includegraphics[width=0.80 \columnwidth]{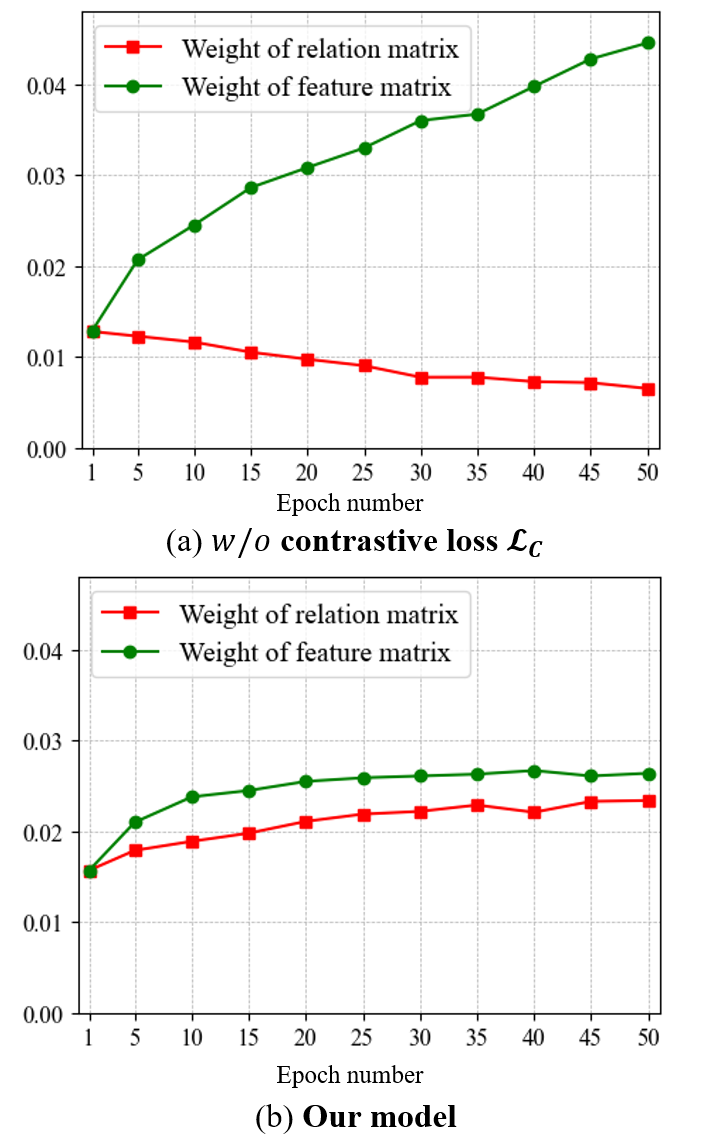} 
\caption{
Illustration of the average weights of feature matrix and relation matrix in the classifier of the inference module.
}
\label{fig5}
\end{figure}

(3) \textit{w/o contrastive loss} $\mathcal{L_C}$.
When we discard the contrastive loss during the second training stage, the performance of our model degrades by \textbf{0.58} $F_1$ points, which confirms that our contrastive loss effectively enhances inference module.
Inspired by \citet{c:172}, we examine  the average weights of the feature matrix and relation matrix in the classifier of the inference module, which can intuitively reflect the importance of the two matrices in relation prediction (See $W_c$ in Equation (\ref{F:12})).
From  Figure~\ref{fig5}(a), we observe that inference module prefers to predict relations using the more stable feature matrix.
However, when introducing the contrastive loss $\mathcal{L_C}$, inference module can simultaneously exploit both feature matrix and relation matrix in relation prediction (See Figure~\ref{fig5}(b)).
Specifically, in this variant, the average weights of the feature matrix and the relation matrix are 0.045 and 0.007, while they are 0.026 and 0.023 in our model.
These results suggest that our contrastive loss can encourage inference module to better exploit the relation matrix.

\begin{table}[]
\centering
\small
\small
{
\begin{tabular}{lcc}
\toprule
Model                                 & Ign$F_1$      & $F_1$     \\ [2pt] \toprule
ATLOP-BERT (our base)              & 59.22       & 61.09          \\ \midrule
Ours-BERT                     & \textbf{60.75}   & \textbf{62.74}         \\ 
\quad w/o fusion sub-layer                      & 60.43       & 62.37       \\
\quad w/o ECA                      & 59.29       & 61.22       \\ 
\quad w/o contrastive loss $\mathcal{L_C}$               & 60.26       & 62.16       \\
\quad w/ negative pairs         & 60.25       & 62.06       \\ 
\quad w/o pre-training                      & 59.51       & 61.45       \\
\quad w/ freeze base module         & 60.35       & 62.31       \\  \bottomrule
\end{tabular}
}
\caption{\label{tab3} Ablation study of our model  on the development set of DocRED.
}
\end{table}

(4) \textit{w/ negative pairs}.
In this variant, we replace our contrastive learning with SimCLR \cite{c:167} and take two different entity pairs from the same entity pair matrix to produce negative pairs.
In addition to pulling together representations of samples in positive pairs like our contrastive learning, this variant also pushs the representations of samples in negative pairs far apart.
Apparently, the performance drop reported in line 7 indicates that pushing entity pairs far apart limits the potential of inference module to capture the dependencies among entity pairs.

\begin{table*}[t]
\centering
\small
{
\begin{tabular}{lcccccc}
\toprule
Model            & \multicolumn{2}{c}{Dev}                  & \multicolumn{2}{c}{Test} \\
                        \cmidrule(r){2-3}  \cmidrule(r){4-5}
                     & Ign$F_1$         & $F_1$       & Ign$F_1$     & $F_1$  \\ [2pt]\toprule
Coref-RoBERTa-large \cite{c:146}        & 57.35       & 59.43              & 57.90      & 60.25       \\
GAIN-RoBERTa-large \cite{c:105}       & 60.87    & 63.09         & 60.31         & 62.76       \\ 
SSAN-RoBERTa-large \cite{c:134}       & 59.40    & 61.42         & 60.25         & 62.08       \\ 
ATLOP-RoBERTa-large \cite{c:112}       & 61.32       & 63.18             & 61.39      & 62.40       \\ 
DocuNet-RoBERTa-large \cite{c:113}     & 62.23       & 64.12       & 62.39      & 64.55  \\
KD-RoBERTa-large \cite{c:159}        & 62.16     & 64.19    & 62.57     & 64.28 \\ \midrule
Ours-RoBERTa-large         & \textbf{62.66}$\pm$0.11 & \textbf{64.58}$\pm$0.13 & \textbf{62.92} & \textbf{64.88*} \\ \bottomrule
\end{tabular}
}
\caption{\label{tab5}
The performance of the RoBERTa-large-based model on the development and test sets of DocRED. 
* denotes significant at $\rho{<}0.01$ with 1,000 bootstrap tests.
We run experiments 5 times with different random seeds and report the mean and standard deviation on the development set.
We save the best checkpoint on the development set and then report the official test scores on the CodaLab scoreboard.
}
\end{table*}

\begin{table}[t]
\centering
\small
{
\begin{tabular}{lccc} 
\toprule
Model                       & Infer-$F_1$      & Pre.         & Recall         \\ [2pt] \toprule
GAIN-GloVe             &40.82          & 32.76     &54.14           \\
SIRE-GloVe             & 42.72         & 34.83     &55.22        \\ \midrule
BERT-RE             & 39.62         & 34.12     &47.23        \\
GAIN-$\rm BERT$            & 46.89         & 38.71     &59.45           \\\midrule
Ours-BERT        &  \textbf{48.75}           &  \textbf{45.02}     & \textbf{53.15}  \\
\quad w/o ECA          &46.91     &41.05    &54.73  \\
\quad w/o Inference module           &46.76     &38.74    &58.96   \\
\bottomrule
\end{tabular}
}
\caption{Infer-$F_1$ points on the dev set of DocRED.
}
\label{tab4}
\end{table}

(5) \textit{w/o pre-training}.
Different from our two-stage training strategy, this variant directly trains our entire model without the pre-training of base module.
From the line 8 of Table~\ref{tab3}, we can observe a drop of \textbf{1.29} $F_1$ points, confirming that the pre-trained base module can provide inference module with better basic information.

(6) \textit{w/ freeze base module}.
To confirm that the performance improvement of our model mainly benefits from inference module, we freeze base module at the second stage of training.
This variant still achieves an $F_1$ score of \textbf{62.31}, which improves the performance of our base model (ATLOP-BERT) by \textbf{1.22} F1 points.
This also implies that our inference module can be used with other types of base modules in a plug-and-play manner.

\subsection{Analysis of Inference Performance}
To further evaluate the inference ability of our model, we follow \citet{c:105,c:155} to report Infer-$F_1$ points in Table~\ref{tab4}, which only considers the relations engaged in the relational inference process. 
For example, if the relational triples ($e_h,r_1,e_o$), ($e_o,r_2,e_t$) and ($e_h,r_3,e_t$) co-occur in the same document, we take them into account in the calculation of Infer-$F_1$ points.

In terms of Infer-$F_1$, our model yields an improvement of \textbf{1.86} points over GAIN. 
Moreover, the performance of our model sharply drops by \textbf{1.84} Infer-$F_1$ points when we replace ECA units with standard multi-head self-attentions.
Meanwhile, without inference module, our model also suffers from performance degradation of \textbf{1.99} Infer-$F_1$ points.
All these results also strongly demonstrate that our inference module can effectively improve the inference ability of the model.

Finally, we introduce a case study in Appendix~\ref{appendix-c} to visually show the effectiveness of our model.

\subsection{RoBERTa-large-based model}
Following some recent studies\cite{c:112,c:113,c:159}, we also report the performance of the RoBERTa-large based models on the DocRED dataset.
From Table~\ref{tab5}, we can observe that our model consistently outperforms all baselines.
Specifically, our model significantly surpasses our base module(ATLOP-RoBERTa-large) by 1.48 $F1$  points (statistical significance $\rho{<}0.01$), and also exceeds KD-RoBERTa-large by 0.6 $F1$ points on the test sets of DocRED.

\section{Related Work}
Early studies on RE mainly focus on sentence-level RE, which predicts the relation between two entities within a single sentence.
In this aspect, many approaches \cite{c:140,c:141,c:145,c:146,c:147,c:148,c:179,c:180} have been proven to be effective in this task.
However, because many relational facts in real applications are expressed by multiple sentences \cite{c:104}, researchers gradually shift their attention to document-level RE.

To this end, researchers have proposed two kinds of methods: Transformer-based and GNN-based methods.
Due to the fact that GNNs can model the dependencies among entities or mentions and have strong inference ability, many researchers explore GNNs for better document-level RE \cite{c:107,c:130,c:139,c:151,c:108,c:106,c:105,c:134}.
Usually, they first construct a document graph, which uses mentions or entities as nodes and leverages heuristic rules and semantic dependencies to build edges.
Then, they perform inference with GNNs on the graph.
For example, \citet{c:106} treat the document graph as a latent variable which can be dynamically induced via structure attention. 
During this process, the induced graph structure can be exploited for better inference in document-level RE.
\citet{c:105} propose a graph aggregation-and-inference network involving a heterogeneous mention-level graph and an entity-level graph.
These two graphs are utilized to model dependencies among mentions and entities, respectively.
Meanwhile, due to the advantage of Transformer \cite{c:160} on implicitly modeling long-distance dependencies, some studies \cite{c:154,c:111,c:112} directly apply pre-trained language models to document-level RE.
\citet{c:112} propose the ATLOP model that features two techniques:
adaptive thresholding \cite{c:176} and localized context pooling.
However, these two types of methods mainly focus on mention- and entity-level information, ignoring the dependencies among entity pairs in same context, which have an important impact on document-level RE.

In contrast, \citet{c:113} and \citet{c:159} exploit the dependencies among entity pairs to facilitate document-level RE, however, ignore the fact that relation prediction difficulties of entity pairs are different.
Furthermore, \citet{c:155} divide entity pairs into intra- and inter-sentence ones, which are considered to have different prediction difficulties.
However, such a division is too simple to accurately distinguish easily- and difficultly-predicted entity pairs.
Unlike these methods only utilizing feature-level information of entity pairs in one pass, we leverage both feature- and relation-level information of entity pairs through iterative inference, allowing our model to capture the dependencies among entity pairs more comprehensively.

Besides, our work is inspired by the mask-predict decoding strategy for non-autoregressive NMT \cite{c:156,c:178}. 
In this work, we adapt this strategy into document-level RE.
To the best of our knowledge, our work is the first attempt to exploit iterative decoding for this task.
Finally, our work is also related to the studies on GNN-based iterative encoding \cite{c:173,c:174,c:177}.
Unlike these studies, we train inference module to directly correct the noisy predictions of base module in a one-pass manner rather than iterative manner, while iterative inference is used only during testing.

\section{Conclusion and Future Work}
In this work, we have proposed a novel document-level RE model with iterative inference, which mainly contains two modules: base module and inference module.
We first use base module to make preliminary predictions on the relations of entity pairs.
Then, via iterative inference, inference module gradually refines the predictions of base module, which is expected to effectively deal with difficultly-predicted entity pairs depending on other pairs.
Experiments on three public document-level RE datasets show that our model significantly outperforms existing competitive baselines.
In future, we attempt to apply our model to other inter-sentence or document-level NLP tasks, such as cross-sentence collective event detection.

\section*{Acknowledgments}
The authors would like to thank the three anonymous reviewers for their comments on this paper. This research was supported in part by the National Natural Science Foundation of China under Grant Nos. 62076211, U1908216, 62276219, and 61573294.

\section*{Limitations}
The limitations of our method mainly include following two aspects:
1) since our method mainly focuses on relational inference, which is rarely required in sentence-level RE, it has low scalability to sentence-level RE.
2) because relational inference is a complex problem, we require a significant amount of relational inference-specific labeled data to effectively train our model.

\bibliography{anthology,custom}
\bibliographystyle{acl_natbib}

\appendix

\section{Hyper-parameters Setting}
\label{appendix-b}
\begin{table*}[t]
\centering
\small
{
\begin{tabular}{lcccc}
\toprule
\multirow{2}*{Hyper-parameter}       & \multicolumn{2}{c}{DocRED}     &\multicolumn{1}{c}{CDR}       &\multicolumn{1}{c}{GDA}  \\
                    \cmidrule(r){2-3}  \cmidrule(r){4-4}      \cmidrule(r){5-5}
     & BERT      & RoBERTa-large   &SciBERT       &SciBERT  \\  [2pt] \toprule
 \multicolumn{5}{c}{\textit{At the first stage}} \\\midrule
Batch size                    &8	    &4	    &8    &6         \\ 
Epoch number                       &30	    &30	 &20   &10 \\
The learning rate for encoder                 &5e-5	  &5e-5  &1e-5	    &2e-5        \\
The learning rate for classifier       &1e-4	  &1e-4  &5e-5	    &8e-5         \\ \bottomrule
 \multicolumn{5}{c}{\textit{At the second stage}} \\\midrule
Batch size                    &8	    &4	    &8    &6         \\ 
Epoch number                       &15	        &15	    &10   &5               \\
The learning rate for base module                 &1e-5	  &5e-6  &1e-6	    &2e-6        \\
The learning rate for Inference module       &1e-4	  &1e-4  &5e-5	    &8e-5         \\ \bottomrule
\end{tabular}}
\caption{\label{tab6} Hyper-parameter Setting.}
\end{table*}

\begin{figure*}[t]
\centering
\includegraphics[width=1.0 \textwidth]{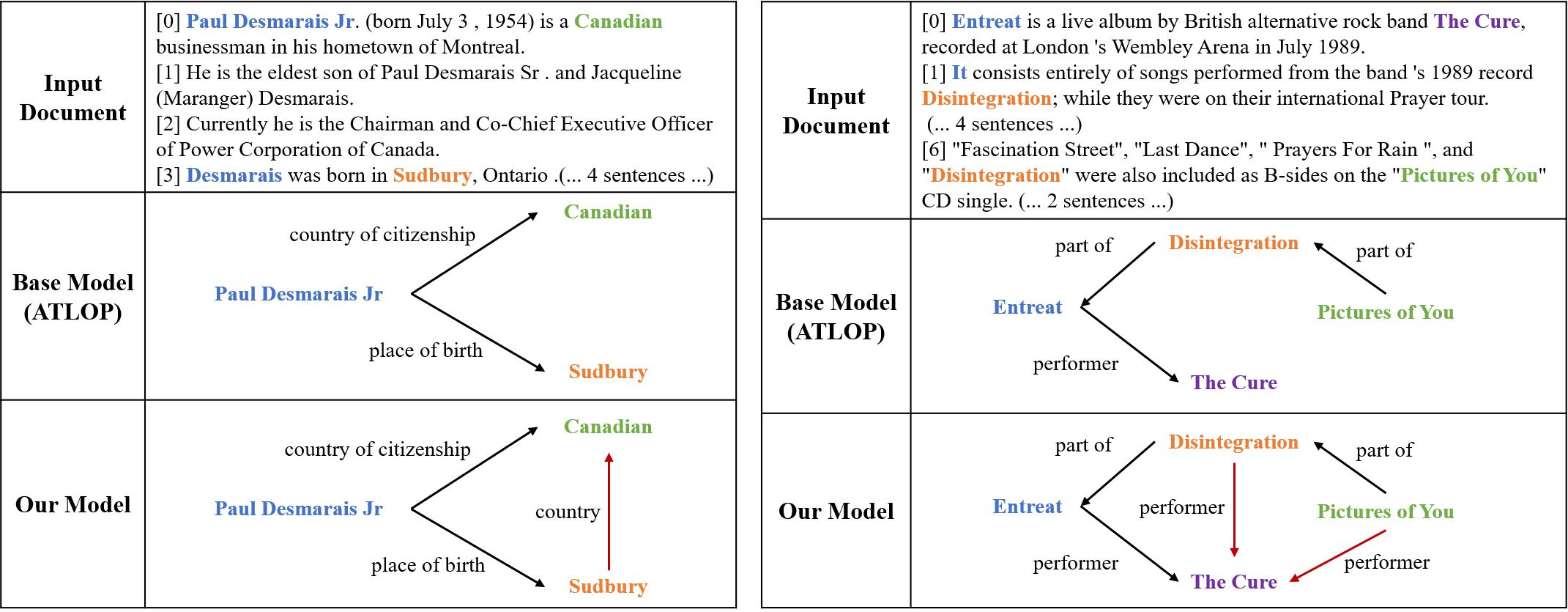} 
\caption{
The case study of our model and our base model (ATLOP). 
We can observe that our base model usually focuses on identifying the relations of easily-predicted entity pairs.
Meanwhile, based on the predictions of base model, our inference module can infer the relations of difficultly-predicted entity pairs.
We only show a part of entities within the documents and the according sentences due to the space limitation.
}
\label{fig6}
\end{figure*}

Table~\ref{tab6} details our hyper-parameter setting. All of our hyper-parameters are tuned on the development set.

\section{Supplementary Experiments}
\label{appendix-c}

\textbf{Case Study}
Figure~\ref{fig6} shows the case study of our model and our base model (ATLOP).
We can observe that ATLOP usually focuses on easily-predicted entity pairs, of which predictions do not require reference to the predicted results of other pairs.
Meanwhile, based on the predictions of base model, our inference module can infer difficultly-predicted entity pairs, of which prediction depends on the predicted results of its overlapping pairs.



\begin{table}[]
\centering
\small
{
\begin{tabular}{lcc} 
\toprule
Model               & CDR          & GDA         \\ [2pt] \toprule
Ours-SciBERT         & \textbf{73.2} & \textbf{85.9}    \\
\quad w/o ECA               & 70.5         & 84.0        \\
\quad w/o fusion sub-layer                & 72.4         & 85.2        \\
\quad w/o contrastive loss                & 71.7         & 84.6        \\
\bottomrule
\end{tabular}
}
\caption{\label{tab7}
Ablation study of our model on the CDR and GDA test sets.
}
\end{table}

\textbf{Ablation studies on CDR and GDA:}
To further illustrate the effectiveness of different modules in the biomedical field, we also conduct ablation experiments on CDR and GDA datasets.
From Table~\ref{tab7}, we observe that all the components contribute to the model performance on these two biomedical datasets.


\end{document}